\documentclass[10pt]{article} %
\usepackage[accepted]{tmlr}
\usepackage[pagebackref,breaklinks,colorlinks]{hyperref}
\usepackage{comment}
\usepackage{pifont}
\usepackage{graphicx}
\usepackage{float}
\usepackage{tikz}
\usepackage{comment}
\usepackage{amsmath,amssymb} %
\usepackage{color}
\usepackage{booktabs}
\usepackage{bbding}
\usepackage{subcaption}
\usepackage{makecell}
\usepackage{color}
\usepackage{multirow}
\usepackage[capitalize]{cleveref}
\usepackage{soul}
\usepackage{colortbl}
\usepackage{algorithm}
\usepackage{listings}
\usepackage{wrapfig}
\usepackage{xspace}

\newcommand{\OURS}{MUG\xspace}

\def\eg{\emph{e.g.}}
\def\ie{\emph{i.e.}}

\definecolor{deepgreen}{rgb}{0,0.5,0}
\definecolor{mygray}{gray}{0.91}
\definecolor{fontgray}{gray}{0.7}

\def\month{August}  %
\def\year{2024} %
\def\openreview{\url{https://openreview.net/forum?id=OCnTRfmaTb}} %

\begin{document}

\title{Vision Learners Meet Web Image-Text Pairs}

\author{Bingchen Zhao$^1$, Quan Cui$^2$, Hao Wu$^2$, Osamu Yoshie$^3$, Cheng Yang$^2$, Oisin Mac Aodha$^1$\\
\addr\small $^1$ University of Edinburgh \qquad $^2$ Bytedance \qquad $^3$ Waseda University
}
\maketitle

\begin{abstract}
Many self-supervised learning methods are pre-trained on the well-curated ImageNet-1k dataset. 
In this work, given the excellent scalability of web data, we consider self-supervised pre-training on noisy web sourced image-text paired data.
First, we conduct a benchmark study of representative self-supervised pre-training methods on large-scale web data in a like-for-like setting. 
We compare a range of methods, including single-modal ones that use masked training objectives and multi-modal ones that use image-text constrastive training. 
We observe that existing multi-modal methods do not outperform their single-modal counterparts on vision transfer learning tasks. 
We derive an information-theoretical view to explain these benchmark results, which provides insight into how to design a novel vision learner. 
Inspired by this insight, we present a new visual representation pre-training method, MUlti-modal Generator~(MUG), that learns from scalable web sourced image-text data.
MUG achieves state-of-the-art transfer performance on a variety of tasks and demonstrates promising scaling properties.  
\end{abstract}
\vspace{-5pt}

\section{Introduction}
\label{sec:intro}

Self-supervised representation learning (SSL) has attracted a considerable amount of attention recently as it offers the potential to reduce the reliance on labor-intensive human collected annotations when training models. 
The design of SSL methods is mainly guided by the InfoMax principle~\citep{infomax}, which shows that maximizing the mutual information $I(X; Z)$ of the input data $X$ and the learned representation $Z$ can lead to better representations. 
In the case of visual data, discriminative contrastive learning~\citep{moco,mocov2,mocov3,simclr,simclrv2,unmix,bardes2022vicreg} and generative masked image modeling~\citep{beit,MAE,baevski2022data2vec,maskfeat,beitv2} have been demonstrated to learn transferable representations from images by attempting to solve pre-defined pretext objectives that aim to indirectly optimize $I(X; Z)$, achieving state-of-the-art results on popular computer vision benchmarks. 
Despite these remarkable advancements, most SSL methods are developed on the ImageNet-1k~\citep{imagenet} dataset, which is well-curated and thus not representative of many real world use cases. 
As a result, these works do not fully exploit the potential scalability of SSL to larger uncurated datasets, \eg, many existing methods are only evaluated on the ImageNet-1k dataset~\citep{xie2021unsupervisedobjectlevel,moco,MAE}. 
There exists works on extending SSL to larger uncurated datasets~\citep{wen2022slotcon,xie2021unsupervisedobjectlevel,detcon,xie2020propagate,wang2020DenseCL}, yet most of these works focus on how to obtain object centric representations~\citep{wen2022slotcon}, or target one specific downstream task, such as object detection~\citep{wang2020DenseCL,xie2020propagate,xie2021unsupervisedobjectlevel,detcon}.

\begin{figure}[t]
  \centering
  \includegraphics[width=0.9\linewidth]{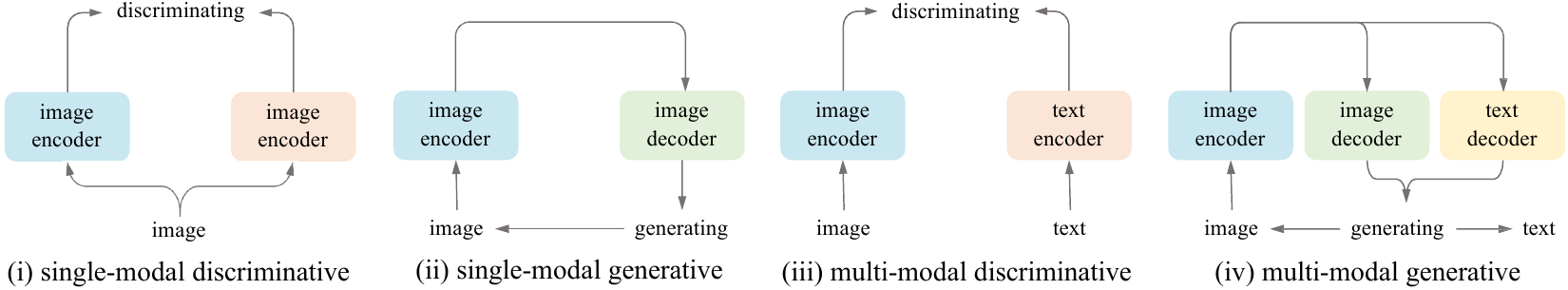}
   \caption{Comparison of different vision pre-training paradigms that use images or image-text pairs. Four paradigms are considered: (i)~single-modal discriminative~(\eg, SimCLR~\citep{simclr}), (ii)~single-modal generative~(\eg, MAE~\citep{MAE}), (iii)~multi-modal discriminative~(\eg, CLIP~\citep{radford2021learning}), and (iv)~our proposed multi-modal generative approach named~\textbf{MUG}. 
   The multi-modal generative paradigm simultaneously generates images and text using only image representations. }
   \label{fig:intro_compare}
\end{figure}

In contrast to the previous single-modal SSL methods, pioneering multi-modal SSL methods~(\eg, CLIP~\citep{radford2021learning} and ALIGN~\citep{align}) have shown impressive scalability due to their ability to train on easily collected large-scale web sourced image-text pairs~\citep{cc3m,thomee2016yfcc100m,laion400m}.  
Multiple works have shown that the textual information present in web image-text pairs can greatly benefit various downstream tasks~\citep{mu2021slip,vilt,filip,declip} and thus contribute to improved transfer learning performance. 
However, many of these existing methods differ not only in the training losses they use, but also in the terms of the model architectures and training data, which makes direct comparison difficult. 
To address this, in this work we conduct a benchmark study of single-modal SSL methods~(\ie, with only images) and multi-modal SSL methods~(\ie, with image-text pairs) using the same web-sourced image-text datasets and training settings.

Based on our benchmark study, we make two key observations. 
First, generative methods achieve the best transfer performance. 
Second, and more surprisingly, the representations learned from multi-modal SSL methods do not outperform single-modal SSL ones.
To explain these observations, we explore an information-theoretical motivated perspective. 
Starting from the InfoMax principle~\citep{infomax} for learning representations, we use the information-bottleneck theory  to explain the transferability of deep representations~\citep{ib1,ib2,vib,ctc}.
We show that generative methods approach a better upper bound on the mutual information $I(X;Z)$, and that introducing multiple generative objectives can further improve upon this bound.

In~\cref{fig:intro_compare}, we group previous representative methods into three families: (i) single-modal discriminative~(\eg, SimCLR~\citep{simclr}), (ii) single-modal generative~(\eg, MAE~\citep{MAE}), and (iii) multi-modal discriminative~(\eg, CLIP~\citep{radford2021learning}). 
Empirical observations and theoretical analyses jointly inspire us to design a fourth paradigm, multi-modal generative pre-training (see~\cref{fig:intro_compare}~\textcolor{red}{(iv)}). 
Under this paradigm, we propose a novel vision learner for web image-text data pre-training, named \textbf{MU}lti-modal \textbf{G}enerator~(MUG).
\OURS is composed of purely multi-modal generative objectives that attempt to generate the  multi-modal joint distribution from a single-modal image input.
From an information-theoretical view, we show that \OURS has a better upper bound for feature transferability, which helps the vision encoder learn more transferable representations when compared to alternative paradigms. 

We present empirical results across a variety of downstream vision tasks to further validate the advantages of \OURS.  
For example, \OURS outperforms the previous best performing methods by $~$2.0\% mIoU when transferred to the ADE20k benchmark and by $~$0.5\% on ImageNet-1k classification. 
Extensive ablations are conducted to help better illuminate the impact of critical model and design choices. 
Finally, visualizations are provided that illustrate that our learned representations are capable of capturing the joint image-text distribution.

\section{Related Work}
\label{sec:related_work}

There is a large body of existing work on the topic of self-supervised learning of representations from visual data. 
We focus on constrastive and masked-based learning objectives, and also discuss learning from paired image and text data. 

\subsection{Vision Learners with Contrastive Learning}
\label{sec:contrastive}
Contrastive learning is a commonly used paradigm for single-modal self-supervised representation learning. 
Here, the pretext task is an instance discrimination one~\citep{inst_disc} which learns a representation by pulling positive examples that are generated from augmentations of the same image closer, while pushing negative examples (\eg, different images from the same mini-batch) apart in the embedding space. 
Building on the basic instance discrimination task, SimCLR~\citep{simclr,simclrv2} showed that larger mini-batches and stronger augmentations can result in more effective representations.
MoCo~\citep{moco,mocov2,mocov3} showed that introducing more negative examples using a momentum encoder and memory bank results in representations that are even more effective for downstream transfer. 
A variety of techniques for improving contrastive learning have been proposed, including hard example mining~\citep{kalantidis2020hard,dvp}, image mixing~\citep{unmix,zhu2021improving}, and by injecting localized priors~\citep{detcon,wen2022slotcon,xie2020propagate,wang2020DenseCL}.

The underlying principle of contrastive learning is the InfoMax principle~\citep{infomax}, which posits that the mutual information $I(X;Z)$ between the input image $X$ and the encoded representation $Z$ should be maximized so that the representation of an image can uniquely identify it from the entire dataset.
CTC~\citep{ctc} showed that the mutual information $I(X;Z)$ is the key to the transferability of one representation, and that preventing the decrease of $I(X;Z)$ during training can improve the transferability of the learned representation.
CTC~\citep{ctc} also showed that contrastive learning objectives are in practice learning to optimize $I(X;Z)=H(Z)-H(X|Z)$ by increasing $H(Z)$.

\begin{equation}
    \underbrace{I(X;Z)}_{\text{Representation Learning}} = \underbrace{H(Z)-H(X|Z)}_{\text{Contrastive Learning}} = \underbrace{H(X)-H(X|Z)}_{\text{Generative Learning}}
\end{equation}

\subsection{Vision Learners with Masked Image Modeling}
\label{sec:mim}
Inspired by the success of masked language modeling (MLM) using transformers~\citep{transformer} for natural language processing~\citep{bert,gpt}, masked {\it image} modeling~\citep{igpt,beit,MAE,xie2021simmim} has been explored in the context of learning representations using vision transformers~\citep{vit,deit}. 
Specifically, iGPT~\citep{igpt} learns a representation by predicting unknown pixels from a sequence of pixels. 
Instead of operating in pixel space, BEiT~\citep{beit} predicts a set of discrete tokens from the encoder of DALLE~\citep{DALLE2,vqvae}, where the tokens are encoded versions of image patches as in~\citep{ramesh2021zero}. 
The training task was further simplified by MAE~\citep{MAE} and SimMIM~\citep{xie2021simmim}, which only enforce the model to reconstruct the masked pixels of the input image using a $\ell_2$ loss and do not require the use of discrete token encoders.

The pretext task of masked image modeling can be viewed as optimizing $-\log p(X|Z)$, where $Z$ is the representation of the non-masked image $X$ and $p(\hat{X}|Z)$ is the likelihood of generating the masked image parts $\hat{X}$ given the learned representation of the input tokens. 
This is equivalent to minimizing $H(X|Z)$. 
Given that $I(X;Z)=H(X)-H(X|Z)$, we can see that masked image modeling still follows the InfoMax~\citep{infomax} principle, where the maximization of $I(X;Z)$ is performed by optimizing $H(X|Z)$, instead of $H(Z)$ as in contrastive learning.
MIM has been shown to be more effective at learning transferable representations compared to contrastive learning~\citep{MAE,xie2021simmim}, indicating the effectiveness of generative pre-training.

\subsection{Vision Learners with Web Image-Text Pairs}
\label{sec:multimodalssl}
The vast majority of the previously discussed works have explored SSL in the context of single-modal well-curated image datasets such as ImageNet-1k~\citep{imagenet}. 
However, it is relatively easy to collect large-scale datasets containing image-text paired data from the web, \eg, images and associated text captions~\citep{cc3m,laion400m}. 
Many approaches have recently began to study what impact this type of data has on learning visual representations, \eg,~\citet{radford2021learning,mu2021slip,vilt,filip,declip,yu2022coca,desai2021virtex}.
CLIP~\citep{radford2021learning} explores this challenge by performing contrastive learning over paired image and text data. 
By leveraging separate visual and text encoders, their training objective aims to maximize the similarity between the learned embeddings of the paired image-text data from each training instance and minimize the similarity between unpaired images and text. 
SLIP~\citep{mu2021slip} builds on this, by learning the representation by jointly performing contrastive learning on both image-text paired datasets like CC3M~\citep{cc3m} and image-only datasets like ImageNet~\citep{imagenet} using strong augmentations~\citep{simclr}.
SLIP showed that the added self-supervised signal on image-only datasets can improve the performance for zero-shot image classification and transfer learning.
However, as we will show later in our like-for-like experiments, the benefit of SLIP over the baseline CLIP model is limited, we conjecture this is due to the fact that the two objectives of SLIP are both contrastive learning losses which are discriminative.  In~\cref{tab:pretrain_comparison}, we show that discriminative objectives are not as effective as generative ones.

CoCa~\citep{yu2022coca} poses representation learning as a cross-modal contrastive learning and auto-regressive caption generation task.  
The contrastive loss of CoCa contrast the image embedding with its corresponding textual caption embedding. 
The caption generation loss is designed to first encode the text only features using a text encoder with causal attention, and then a transformer decoder with cross-attention to image encoded features are used to decode the caption auto-regressively.
These two objectives both enforce the model to learn a predictive image representation of the caption. 
CapPa~\citep{tschannen2023image} further explores visual representation learning using only an auto-regressive image captioning loss. 
High performance of zero-shot learning and few-shot linear classification is achieve in CapPa.
In contrast, in our designed model in~\cref{sec:approach}, we enforce the image representation to not only be predictive of the caption, but also to be predictive of the image itself.
Additionally, this paper also provides an intuitive understanding of why and how such generative training objective results in a strong representation learning ability from an information theoretical perspective.

There also exists other work that focuses on achieving higher zero-shot performance from a frozen visual representation~\citep{zhai2022lit} by filtering the caption to achieve better performance~\citep{li2022blip}, or adapting frozen visual and language models together to achieve in-context learning abilities on complex visual-language tasks~\citep{alayrac2022flamingo,li2023blip,koh2023grounding,zong2023self}.
In this work, we first comprehensively study existing paradigms for learning transferable representations from publicly available image-text paired datasets. 
Inspired by this, we propose a new method for learning representations and further compare our method to the previous state-of-the-art works where we out perform them across a wide variety of tasks.

\section{Preliminaries}
\label{sec:pre}

\subsection{Benchmarking SSL Methods on Web Data}
Given a dataset of image-text pairs $\mathcal{D} = \{(x_i^V,x_i^L)\}$ sampled from the joint distribution of $p(X^V, X^L)$, where an image is of size $x_i^V\in \mathbb{R}^{C\times H\times W}$, and a piece of text is of size $x_i^L\in \mathbb{R}^{L}$, our goal is to learn a visual feature encoder $f: \mathbb{R}^{C\times H\times W}\rightarrow \mathbb{R}^D$ that maps an input image into a compact feature embedding and enables effective transfer learning performance~\citep{rethinkingpretrain1}.

Two families of self-supervised learning methods can be used to achieve this goal. 
This first is single-modal SSL, which uses only the images $x_i^V$ within the dataset $\mathcal{D}$. 
\cref{sec:contrastive} and \cref{sec:mim} introduce and describe the current most representative methods for single-modal SSL.
The second approach is multi-modal SSL. 
This aims to learn an effective representation by aligning the representation of the input image  with the corresponding paired input text. 
A brief introduction to multi-modal SSL methods is provided in \cref{sec:multimodalssl}.
As a preliminary motivation for our work, we first perform a benchmark analysis. 
We pre-train representative SSL methods on the \textit{same} web image-text dataset, for the \textit{same} number of training epochs. 
Then, we evaluate the transfer performance on downstream tasks by fine-tuning the resulting pre-trained models.

\noindent \textbf{Experimental setup.}
We categorize methods according to their pre-training paradigm, as discussed in~\cref{fig:intro_compare}. 
For single-modal SSL methods, we choose MoCoV3~\citep{mocov3} and SimCLR~\citep{simclr} as representative discriminative methods, and MAE~\citep{MAE} as a representative generative method.
For image-text multi-modal SSL, we choose CLIP~\citep{radford2021learning} and SLIP~\citep{mu2021slip} as discriminative methods, and also include CoCa~\citep{yu2022coca}, which is composed of both discriminative and generative targets. 
Notably, in addition to these methods, we further implement another multi-modal baseline namely Masked Captioner~(MAC), which generates text from masked images.
It is trained using only a purely generative cross-modal objective. 
The details of MAC will be introduced in~\cref{sec:approach}.
For the pre-training hyper-parameters, we strictly follow the recommended setting in the original paper of each method. 
For a fair comparison, we pre-train all methods on a web dataset CC3M~\citep{cc3m} for 400 epochs. 
When transferring the representation to ImageNet-1k~\citep{imagenet}, we follow the widely used fine-tuning recipe introduced by~\citet{beit,MAE}.

\begin{figure*}[t]
  \centering
  \includegraphics[width=1.0\linewidth]{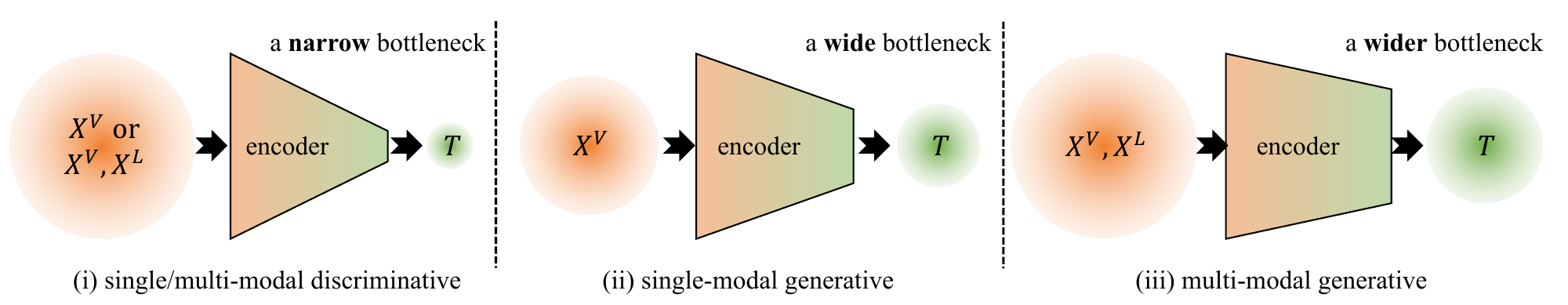}
   \caption{\textbf{\textit{Left:}} Single/multi-modal discriminative methods have a narrow bottleneck and thus learn a less informative representation. \textbf{\textit{Middle:}} Single-modal generative methods have a wide bottleneck and thus learn a more informative representation. \textbf{\textit{Right:}} Multi-modal generative methods have a wider bottleneck for generating~(\eg, recovering) the joint distribution of both modalities, and as a result learn an even more informative representation.}
   \label{fig:local_align}
\end{figure*}

\begin{table}[t]
    \centering
    \small
    \caption{Like-for-like comparison of existing SSL pre-training methods. All models are pre-trained on the CC3M dataset~\citep{cc3m}, and evaluated on ImageNet-1k~(IN-1K)~\citep{imagenet} via end-to-end fine-tuning. ``gen. and dist.'' are short hand for generative and discriminative. ``$\mathcal{I}$ and $\mathcal{T}$'' denote images and text. For instance, SimCLR~\citep{simclr} is pre-trained by discriminating images~(disc. $\mathcal{I}$). CLIP~\citep{radford2021learning} is pre-trained by discriminating image-text pairs~(disc. $\mathcal{I}$\&$\mathcal{T}$). CoCa~\citep{yu2022coca} has two objectives: discriminating image-text pairs~(disc. $\mathcal{I}$\&$\mathcal{T}$) and generating text from images~(gen. $\mathcal{T}$ w. $\mathcal{I}$). We also include a  MAsked Captioner~(MAC) baseline that generates text from masked image patches~(gen. $\mathcal{T}$ w. $\mathcal{I}$). 
    }
    \label{tab:pretrain_comparison}
    \begin{tabular}{llc}
    \toprule
    method          & pre-training paradigm          &  IN-1K~\citep{imagenet}  \\
    \midrule
    \multicolumn{3}{l}{\textit{single-modal pre-training}}   \\
    MAE~\citep{MAE}  &  \textbf{gen.} $\mathcal{I}$        &    \textbf{83.0}      \\
    SimCLR~\citep{simclr}  &  disc. $\mathcal{I}$   &    82.7      \\
    MoCoV3~\citep{mocov3} &  disc. $\mathcal{I}$    &    82.6      \\
    \midrule
    \multicolumn{3}{l}{\textit{multi-modal pre-training}}   \\
    MAC &  \textbf{gen.} $\mathcal{T}$ w. $\mathcal{I}$  &       \textbf{81.7}    \\
    CLIP~\citep{radford2021learning} &  disc. $\mathcal{I}$\&$\mathcal{T}$  &  79.7      \\
    SLIP~\citep{mu2021slip} &  disc. $\mathcal{I}$\&$\mathcal{T}$ + disc. $\mathcal{I}$  &       80.9    \\
    CoCa~\citep{yu2022coca} &  disc. $\mathcal{I}$\&$\mathcal{T}$ + gen. $\mathcal{T}$ w. $\mathcal{I}$  &       79.5    \\
    \bottomrule
    \end{tabular}
\end{table}

\subsection{Key Observations}
\label{sec:discussion}

We make the following key observations from the results of the experiments in~\cref{tab:pretrain_comparison}:

\noindent \textbf{Generative methods~(\eg, MAE) achieve the best results.} Across all methods, we can observe that MAE~\citep{MAE} achieves the best performance. 
As mentioned in~\cref{sec:contrastive} and~\cref{sec:mim}, we argue that discriminative pre-training and generative pre-training approach the InfoMax~\citep{infomax} principle via different mechanisms. 
Specifically, discriminative pre-training optimizes $I(X;Z_d)=H(Z_d)-H(Z_d|X)$ via maximizing $H(Z_d)$, while generative pre-training optimizes $I(X;Z_g)=H(X)-H(X|Z_g)$ via minimizing $H(X|Z_g)$~\citep{ctc}, here $Z_d$ is the representation learned by discriminative pre-training, and $Z_g$ is the representation learned by generative pre-training.
Suppose the models learned by generative pre-training and discriminative pre-training both achieve zero training loss, then $I(X;Z_d)$ will be $H(Z_d)$ and $I(X;Z_g)$ will be $H(X)$. In the discriminative pre-training scenario, if we consider the Markov Chain of $X\rightarrow Z_d$, we have $I(X;X) \geq I(X;Z_d)$ from the data process inequality, \ie, $H(X) \geq I(X;Z_d)=H(Z_d)$. 
Thus generative pre-training has a better transferability upper bound compared to discriminative pre-training, and results in more informative and beneficial features for downstream tasks.

\begin{align}
    \text{Disc.: }I(X;Z_d)=\underbrace{H(Z_d)}_{\text{maximizing}}-H(Z_d|X) \xrightarrow{\text{upperbound}} &H(Z_d)  \\
    \text{Gen.: }I(X;Z_g)=H(X)-\underbrace{H(X|Z_g)}_{\text{minimizing}} \xrightarrow{\text{zero training loss}} &H(X)
\end{align}

\noindent \textbf{Current multi-modal methods do not outperform single-modal ones.} We can observe that the best single-modal pre-training method outperforms the best multi-modal pre-training method by $1.3$\% on the ImageNet-1k accuracy. 
Compared with our introduced MAC baseline, multi-modal discriminative methods (\eg, CLIP and CoCa) yield worse results. We believe the reason is that \textit{the information from the text modality is greatly limited}. 
A typical text annotation only contains partial information about the contents of the corresponding image. 
A prominent researcher has commented that ``language is a very low-bandwidth method for transmitting information: isolated words or sentences, shorn of context, convey little.''~\citep{browning2022ai} It suggests the $H(X^L)$ could be much lower than $H(X^V)$. 
In the case of the multi-modal discriminative method CLIP, a low $H(X^L)$ directly limits $I(X^L;Z_d)$. 
Thus, poor transferring performance is observed. 
When comparing SimCLR and SLIP we observe that SLIP falls behind SimCLR by $1.8$\% accuracy, which further supports the potential limitations of multi-modal discriminative training objectives. 
However, MAC is generative and alleviates this problem, but it still cannot outperform single-modal methods.

\begin{equation}
    H(X^L) \ll H(X^V)
\end{equation}

\noindent \textbf{Discussions and opportunities.} 
Inspired by the information-bottleneck theory~\citep{ib1,ib2,vib,ctc,infomax}, we illustrate to help understand our observations in~\cref{fig:local_align}.
The first observation is that generative methods can help the model learn more transferable visual representations compared to discriminative ones due to a superior $I(X;Z)$ upper bound. In~\cref{fig:local_align}~(left), we argue that discriminative methods have a narrow bottleneck and a low $I(X;Z)$, thus a less informative representation learned. More detailed discussions are provided in Appendix~\cref{app:sec:discussion}.

In~\cref{fig:local_align}~(middle), generative methods have a wider bottleneck, leading to more informative representations. 
Given the side effects of multi-modal discriminative objectives~(the second observation), developing multiple generative objectives will encourage the model to learn more informative representations. As shown in~\cref{fig:local_align}~(right), for an image-text pair, generating the joint distribution of both the image and text modality will help the model to ``absorb'' as much information as possible from the data and approach a larger $I(X;Z)$.

Note that here we are using the term bottleneck to refer to the variance in the representation space $Z$ with respect to the input images $X$, \ie, how much the representation changes when the input is changed.
One way to understand this bottleneck is from the perspective of the rank of the feature matrix~\citep{garrido2022rankme}. 
A higher rank would indicates that the model can capture more variance in the input, whereas a collapsed lower rank captures less information about the input. 
In contrast, a completely collapsed representation would map all inputs to a same feature which has a rank of zero.
In~\cref{sec:rankme}, we provide a qualitative measure of the rank of different pre-trained models. %

\section{Approach}
\label{sec:approach}

\subsection{Motivation}
Following the above observations, in this section we describe a more effective vision learner that learns from web sourced image-text paired training data. 
We first formulate the mechanism that a multi-modal generative objective uses to learn a more transferable representation, in line with~\cref{fig:local_align}~(right). 
We believe that introducing multiple generative objectives is feasible, where now the optimization objective is to learn how to generate the joint distribution $p(X^V, X^L)$ from a single masked input distribution $\hat{X}^V$ as our focus is on learning a visual encoder. 

To generate the joint distribution $p(X^V, X^L)$ from the latent representation $Z=f(\hat{X}^V)$ extracted from the non-masked visual input $\hat{X}^V$, the model maximizes the following mutual information:
\begin{align}
    I(X^V, X^L;Z)= & H(X^V, X^L) - H(X^V, X^L|Z) \nonumber\\
                 \geq &~\text{max}\left[H(X^V), H(X^L)\right] - H(X^V, X^L|Z) \\
                 \rightarrow &~\text{max}\left[H(X^V), H(X^L)\right] \text{(zero training loss)}
\end{align}
If we consider the fully optimized model to the optimal, then $H(X^V, X^L|Z)=0$. 
As the lower bound of $I(X^V, X^L;Z)$ is greater than or equal to the lower bound of $I(X^V;Z)$ because of the added variable $X^L$, we can observe that generating the joint distribution of $p(X^V, X^L)$ can at least learn an equally transferable representation as single-modal generative pre-training. In practice, we observe that multi-modal generative pre-training achieves superior transferability.
This design motivation is also in line with the discussion in~\cref{sec:discussion}. 
Overall this intuition motivates us to design a generative pre-training framework with multiple generation tasks using only the visual input and representation.

\begin{figure}[!t]
\centering
\begin{minipage}{1.0\linewidth}
\centering
\begin{minipage}{0.50\linewidth}
\begin{figure}[H]
  \centering
  \includegraphics[width=\linewidth]{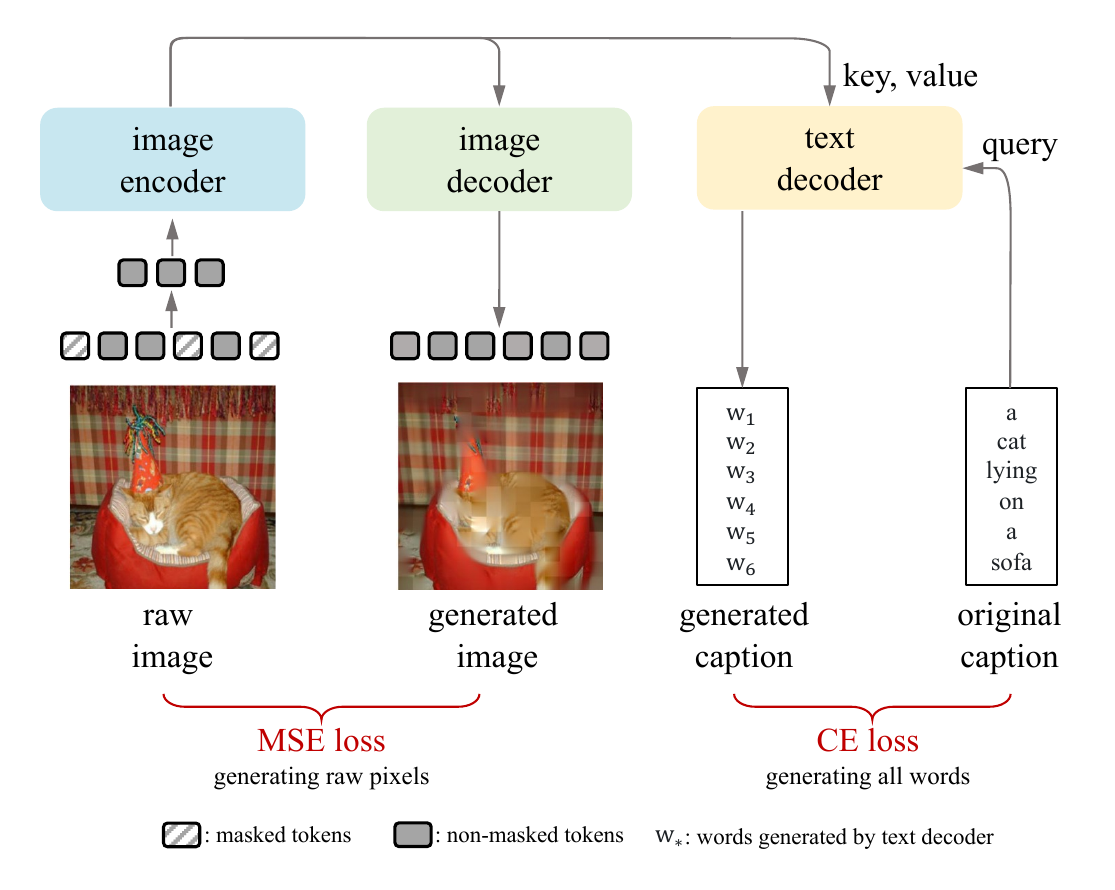}
   \caption{Illustration of our multi-modal self-supervised approach \OURS. 
   }
   \label{fig:mug_framework}
\end{figure}
\end{minipage}
\hspace{0.03\linewidth}
\begin{minipage}{0.45\linewidth}
\begin{algorithm}[H]
\caption{Pseudocode for \OURS. 
}
\label{alg:code}
\definecolor{codeblue}{rgb}{0.25,0.5,0.5}
\lstset{
  backgroundcolor=\color{white},
  basicstyle=\fontsize{7.2pt}{7.2pt}\ttfamily\selectfont,
  columns=fullflexible,
  breaklines=true,
  captionpos=b,
  commentstyle=\fontsize{7.2pt}{7.2pt}\color{codeblue},
  keywordstyle=\fontsize{7.2pt}{7.2pt},
}
\begin{lstlisting}[language=python]
# img, txt: image-text paired data
# txt_mask: causal mask for captioning
def text_decoder(q, kv, mask):
    q = single_modal_attn(q, mask)
    cap_res = multimodal_attn(q, kv)
    return cap_res

patch_img = patchify(img)
masked_token = masking(patch_img) # [N, L, D]
latent = vit_encoder(masked_token) # [N, L, D]

# Generative objective for image
recon_img = mae_decoder(latent) # [N, L, D]
recon_loss = mse_loss(img, recon_img)

# Generative objective for text
label, txt = txt[:, 1:, :], txt[:, :-1, :]
txt_feat = tokenizer(txt) # [N, L, D]
cap_res = text_decoder(q=txt_feat, kv=latent, mask=txt_mask)
cap_loss = ce_loss(label, cap_res)

loss = recon_weight * recon_loss + cap_weight * cap_loss
\end{lstlisting}
\end{algorithm}
\end{minipage}
\end{minipage}
\end{figure}

\subsection{MUG: MUlti-modal Generator}
Here we describe our novel multi-modal generative vision learner, named MUlti-modal Generator~(MUG). The framework of MUG is presented in~\cref{fig:mug_framework}. 
It consists of an image encoder, an image decoder, and a text decoder. %
Two tasks are involved, \ie, raw pixel and original caption generation.

\noindent \textbf{Raw pixel generation.}
Given a masked visual input $\hat{x}_i^V=M\odot x_i^V$, where $M$ is the random generated $0$-$1$ mask following the procedure in MAE~\citep{MAE} where $0$ indicates that the image region is masked out and $1$ indicates that it is visible to the model, $\odot$ denotes the element-wise product, the visual feature encoder $f$ maps the masked input image to a latent representation $t_i^V=f(\hat{x}_i^V)$. 
The latent representation $t_i^V$ is then fed into a visual decoder $g_V$ to generate $\tilde{x}_i^V=g_V(t_i^V)$ which is the reconstruction of the non-masked visual input. We adopt the reconstruction loss from MAE~\citep{MAE} to train both $f$ and $g_V$:
\begin{equation}\label{eq:mae}
    \mathcal{L}_{V}=\frac{1}{\Omega (\hat{x}_i^V)} \|(1-M)\odot x_i^V - (1-M)\odot \tilde{x}_i^V\|^2,
\end{equation}
where $\Omega(\hat{x}_i^V)$ is the number of masked elements in $\hat{x}_i^V$. This loss encourages the generation of the visual part of the input.

\noindent \textbf{Caption generation.}
We further propose to generate the paired text (\ie, the image caption) $x_i^L$ from the feature $t_i^V$ of the masked visual input $\hat{x}_i^V$.
To do this, we feed the feature $t_i^V$ to a text decoder $g_L$ to generate the text captions $\tilde{x}_{i,j}^L=g_L(x_{i,<j}^L,t_i^V)$ in an auto-regressive fashion, driven by the following loss:
\begin{equation}
    \mathcal{L}_{L}=-\sum_{j=1}^{J}\log P(\tilde{x}_{i,j}^L | x_{i, <j}^L, t_i^V),
\end{equation}
where $J$ is the maximum length of a textual caption, $x_{i,<j}^L$ is the first $j-1$ tokens of the caption and $x_{i,j}^L$ is the $j$-th token of the caption. 
The text generation is conditioned on the image features, which helps to improve the transferability of the image encoder. 
Thus, the proposed text decoder has two parts, \ie, single-modal and multi-modal layers. 
The single-modal part is constructed as an auto-regressive model via a causal mask~\citep{transformer}, where each word can only attend to the words before itself in a caption sequence. 
It only provides the text decoder with contextual information when generating the caption.
We design the multi-modal part using the image features as the \textit{key} and \textit{value} and the output from single-modal layers as the \textit{query} in a cross-attention layer~\citep{transformer}.
This ensures the text generation is derived from the image features as the contextual information only serves as the \textit{query}.
Thus, the entire text decoder can be seen as regrouping the image features $t_i^V$ with contextual information to generate captions.

\noindent \textbf{Optimization.}
Our multi-modal pre-training framework is trained by optimizing the above two losses:
\begin{equation}\label{eq:overall}
    \mathcal{L} = \lambda_{V} \mathcal{L}_{V} + \lambda_{L} \mathcal{L}_{L},
\end{equation}
where $\lambda_{V}$ and $\lambda_{L}$ are different weights for the image and text generation losses.
Additionally, for the MAC method we evaluate in~\cref{sec:pre}, only the caption loss $\mathcal{L}_{L}$ is used to optimize the model.

\subsection{Understanding the Loss Function}

If we consider the \OURS image decoder as modeling the image using a Gaussian prior~\citep{zhao2015loss}, then the loss in~\cref{eq:mae} is actually maximizing the log-likelihood of the generated image:
\begin{align}
    \log P_{x_i^V}(\tilde{x}_i^V)&=\log\frac{1}{\sigma \sqrt{2\pi}}\exp\left(-\frac{1}{2}\left(\frac{\tilde{x}_i^V-x_i^V}{\sigma}\right)^2\right) =-\frac{1}{\mathrm{M}}\|\tilde{x}_i^V-x_i^V\|^2-\mathrm{C},
\end{align}
where $\sigma$ is the standard deviation of the distribution. $\mathrm{M}$ and $\mathrm{C}$ are constants that do not affect the optimization process.
Together with $\mathcal{L}_{L}$, we can see that our loss function in~\cref{eq:overall} is maximizing the log-likelihood of both the generated image and the generated paired text.
Thus the latent representation $Z$ extracted by the model is optimized to be more predictive of the image $X^V$ and the caption $X^L$. In other words, the conditional entropy $H(X^V,X^L|Z)$ is minimized via the two losses in our framework, maximizing the mutual information $I(X^V,X^L;Z)$, and leading to a more transferable representation.

\begin{table}[t]
    \centering
    \small
    \caption{Comparison of different methods pre-trained on CC3M~\citep{cc3m} and  then evaluated via fine-tuning on ImageNet-1k~(IN-1K)~\citep{imagenet}. Top-1 accuracy results are reported. In addition, we evaluate the IN-1k fine-tuned models on ImageNet-Adversarial~(IN-A)~\citep{imagenet-a} and ImageNet-Rendition~(IN-R)~\citep{imagenet-r} to assess out-of-distribution performance. 
    }
    \label{tab:in1k_ft_main_result}
    \setlength{\tabcolsep}{4pt}
    \begin{tabular}{lccccc}
    \toprule
    method   & data  & \#epoch     & IN-1K & IN-A         &  IN-R  \\
    \midrule
    \multicolumn{4}{l}{\textit{single-modal pre-training}} \\
    SimCLR  & CC3M & 400  & 82.7 &  31.2        &    49.1      \\
    MoCoV3  & CC3M & 400  & 82.6 &  29.7        &    46.9      \\
    MAE     & CC3M &  400  & 83.0 &  32.9        &    49.4      \\
    \midrule
    \multicolumn{4}{l}{\textit{multi-modal pre-training}} \\
    CLIP  & CC3M & 400  & 79.7 &  23.6        &    42.6      \\
    SLIP  & CC3M & 400  & 80.5 &  20.5        &    44.4      \\
    CoCa  & CC3M & 400  & 79.5 &  18.1        &    42.9      \\
    MAC~(impl.)  & CC3M & 400  & 81.7 &  27.5        &    44.3      \\
    \rowcolor{mygray} \OURS~(ours)  & CC3M & 400  & \textbf{83.5} &  \textbf{36.3}        &    \textbf{50.4}       \\
    \bottomrule
    \end{tabular}
\end{table}

\vspace{-.3cm}
\section{Experiments}
\label{sec:experiment}
\vspace{-.3cm}

\subsection{Implementation Details}

\noindent \textbf{Encoders and decoders.}
Our framework contains an image encoder $f$, and two decoders $g_V$ and $g_L$ for decoding the latent representation to reconstruct the image and the textual caption, respectively.
The image encoder $f$ is implemented using a Vision Transformer~\citep{vit} backbone, and the image decoder is implemented with the same architecture as in MAE~\citep{MAE}. 
We implement the text decoder $g_L$ following SimVLM~\citep{simvlm}. 
For efficient training, we train our encoder-decoder architecture using teacher-forcing~\citep{teacherforcing}.
Detailed implementation details, including architectures and hyper parameters, are provided in the Appendix. 
The code is available at~\url{https://huggingface.co/spaces/tennant/MUG_caption}.

\noindent \textbf{Pre-training datasets.}
We train models on the publicly available CC3M~\citep{cc3m} and LAION400M~\citep{laion400m} datasets. 
Additionally, we also collect 200M web image-text pairs to study the scalability of \OURS. 
The collected data has no filtering processing and as a result it potentially contains a large amount of noise which presents a challenge for multi-modal methods. 
We denote this privately collected dataset as W200M.

\subsection{Transfer Learning}

We begin by evaluating the transfer performance of different  learned representations on a wide range of downstream tasks. 
Transfer learning is performed with conventional full end-to-end fine-tuning of the model weights on target task~\citep{MAE}.
This highlights the superiority of \OURS compared to other methods.

\noindent \textbf{Image classification on ImageNet-1k.}
First, we transfer learned representations to general image classification on ImageNet-1k~\citep{imagenet} using end-to-end fine-tuning. 
In addition to reporting performance on the validation images in ImageNet-1k, we also evaluate ImageNet-1k fine-tuned models on out-of-distribution datasets ImageNet-Adversarial~(IN-A)~\citep{imagenet-a} and ImageNet-Rendition~(IN-R)~\citep{imagenet-r}.
Results are reported in~\cref{tab:in1k_ft_main_result} and~\cref{tab:in1k_ft_backbone}. 
Compared to previous methods, \OURS achieves the best top-1 accuracy 83.5\%~(with CC3M).%
When training \OURS with ViT-S, ViT-B, and ViT-L backbones, we also observe consistent improvements  over the previous single-modal best performing MAE~\citep{MAE}. 
Both comparisons illustrates that \OURS surpasses previous methods by making use of scalable web image-text paired data.
Notably for the results on IN-A and IN-R, \OURS outperforms the previous multi-modal methods by a large margin (\ie, $>$10\% on IN-A)  even through our model leverages the same supervision signal as these other multi-modal methods. 
This indicates that \OURS is much more effective for learning representations that generalize to hard out-of-distribution cases.

\begin{wraptable}[17]{l}{.45\linewidth}
    
    \centering
    \small
    \caption{Comparison of different backbone sizes (\ie, ViT-S, ViT-B, and ViT-L) for models pre-trained with CC3M for 400 epochs and then fine-tuned on IN-1K. Top-1 accuracy is reported. Across all backbone sizes, we observe that \OURS is superior to MAE in almost all cases. }
    \label{tab:in1k_ft_backbone}
    \begin{tabular}{lcccc}
    \toprule
    method     & backbone     & IN-1K & IN-A         &  IN-R  \\
    \midrule
    MAE  & ViT-S  & 80.9 &  22.7        &    \textbf{44.1}      \\
    \rowcolor{mygray} \OURS~(ours)  & ViT-S  & \textbf{81.6} &  \textbf{25.0}        &    42.5      \\
    \midrule
    MAE  & ViT-B  & 83.0 &  32.9        &    49.4      \\
    \rowcolor{mygray} \OURS~(ours)  & ViT-B  & \textbf{83.5} &  \textbf{36.3}        &    \textbf{50.4}      \\
    \midrule
    MAE  & ViT-L  & 85.0 &  49.4        &    59.6      \\
    \rowcolor{mygray} \OURS~(ours)  & ViT-L  & \textbf{85.4} &  \textbf{53.8}        &    \textbf{63.1}      \\
    \bottomrule
    \end{tabular}
\end{wraptable}

Next, we evaluate pre-trained models using  linear probing evaluation on ImageNet-1k. Results are reported in~\cref{tab:in1k_lp_main_result} and~\cref{tab:in1k_lp_backbone}. 
Compared with discriminative pre-training methods, most generative methods obtain unsatisfactory IN-1k linear probing results. 
\OURS also suffers from unsatisfactory linear probing performance common to other masked transformer models~\citep{MAE}, where it has been hypothesized that linear probing requires discriminative representations, which cannot be satisfied by generative methods.
For instance, SimCLR and \OURS achieve 63.3\% and 61.3\% on the IN-1k validation set, respectively.
However, in the case of IN-A and IN-R, generative methods like \OURS achieve the best results. 
We conjecture this is due to the fact that discriminative methods can easily be overfit to the in-domain data, but conversely  cannot achieve good out-of-distribution generalization.
Additionally, compared with the original MAE, \OURS outperforms it by a large margin.
Notably, when using  the backbone ViT-S in~\cref{tab:in1k_lp_backbone}, our method outperforms MAE by \textbf{11.3\%} top-1 accuracy. 
This indicates that MAE pre-trained models can suffer from poor results due lower model capacity, whereas MUG pre-trained models can alleviate this drawback with the help of external data information from the text modality.

\vspace{10pt}

\begin{table}[ht]
\parbox{.45\linewidth}{
    \centering
    \small
    \setlength{\tabcolsep}{2pt}
    \caption{
    Comparison of pre-training methods on ImageNet-1k~(IN-1K) using linear probing. Top-1 accuracy is reported.
    }
    \label{tab:in1k_lp_main_result}
    \begin{tabular}{lccccc}
    \toprule
    method   & data  & \#epoch     & IN-1K & IN-A         &  IN-R  \\
    \midrule
    \multicolumn{4}{l}{\textit{single-modal pre-training}} \\
    SimCLR       & CC3M & 400  & \textbf{63.3} &  12.5        &    23.0      \\
    MAE          & CC3M & 400  & 57.4          &  9.1         &    17.8      \\
    \midrule
    \multicolumn{4}{l}{\textit{multi-modal pre-training}} \\
    CLIP         & CC3M & 400  & 55.1          &  11.2        &    18.9      \\
    CoCa         & CC3M & 400  & 56.9          &  13.8        &    20.3      \\
    MAC~(impl.)  & CC3M & 400  & 59.3          &  15.2        &    21.3      \\
    \rowcolor{mygray} \OURS~(ours)  & CC3M & 400  & 61.3 &  \textbf{16.7}        &    \textbf{24.2}       \\
    \bottomrule
    \end{tabular}
}
\hspace{0.05\linewidth}
\parbox{.45\linewidth}{
    \centering
    \small
    \caption{Comparison of pre-training methods when fine-tuning on fine-grained iNat-17, iNat-18, and P365. Top-1 accuracy is reported. 
    \textcolor{fontgray}{MAE$*$ is pre-trained on IN-1K.}
    }
    \label{tab:fg_ft_main_result}
    \begin{tabular}{lcccc}
    \toprule
    method     & backbone     & iNat-17 & iNat-18         &  P365  \\
    \midrule
    MAE  & ViT-S  & 65.4 &  70.0        &    57.9      \\
    \rowcolor{mygray} \OURS~(ours)  & ViT-S  & \textbf{65.5} &  \textbf{70.7}        &    \textbf{58.3} \\
    \midrule
    \color{fontgray}MAE$*$  & \color{fontgray}ViT-B  & \color{fontgray}70.5 &  \color{fontgray}75.4        &    \color{fontgray}57.9      \\
    MAE  & ViT-B  & 67.9 &  73.0        &    58.3      \\
    \rowcolor{mygray} \OURS~(ours)  & ViT-B  & \textbf{68.4} &  \textbf{73.4}        &    \textbf{58.6}      \\
    \midrule
    MAE  & ViT-L  & 73.0 &  76.8        &    59.4      \\
    \rowcolor{mygray} \OURS~(ours)  & ViT-L  & \textbf{75.1} &  \textbf{77.8}        &    \textbf{59.8}      \\
    \bottomrule
    \end{tabular}
}
\end{table}

\begin{wraptable}[9]{r}{0.45\linewidth}
    \centering
    \small
    \caption{Comparison of different backbone sizes (\ie, ViT-S, ViT-B, and ViT-L). Models are pre-trained on CC3M for 400 epochs and evaluated using linear probing on IN-1K.  }
    \label{tab:in1k_lp_backbone}
    \begin{tabular}{lccc}
    \toprule
    method    &  ViT-S  &  ViT-B  & ViT-L \\
    \midrule
    MAE                             &  41.0   &  57.4   &  63.5     \\
    \rowcolor{mygray} \OURS~(ours)                      &  \textbf{52.3}   &  \textbf{61.3}   &  \textbf{67.6}     \\
    \bottomrule
    \end{tabular}
\end{wraptable}

\noindent \textbf{Fine-grained image classification.}
To illustrate that our pre-trained models do not over-fit on ImageNet-1k-based evaluations, we further transfer learned representations to fine-grained image datasets. 
Specifically, we report performance on iNaturalist-17~(iNat-17)~\citep{inat}, iNaturalist-18~(iNat-18), and Places365-Small~(P365)~\citep{places} with end-to-end fine-tuning. 
Results are presented in~\cref{tab:fg_ft_main_result}. 
Similar to the results of ImageNet-1k, \OURS still outperforms the previous SOTA method MAE on all evaluated cases. 
This further validates the success and robustness of \OURS in transferring to image classification tasks. Moreover, we notice that the performance of \OURS is relevant to the pre-training data domain. 
Concretely, iNat-17 and iNat-18 share a similar domain with IN-1K. 
Thus, the IN-1K MAE models perform better than CC3M MAE models. 
P365 shares a similar domain with CC3M, which leads to better results of CC3M-based models. Regardless of the fine-tuning data domain, the improvements from \OURS are stable and generalize across datasets.

\begin{table}[t]
    \centering
    \small
    \caption{Semantic segmentation fine-tuned results (reported as mIoU) comparing different pre-training methods on ADE20K~\citep{ade20k}.
    \textcolor{fontgray}{MoCoV3$*$ and MAE$*$ are pre-trained on IN-1K.}}
    \label{tab:ade20k_ft_main_result}
    \begin{tabular}{lcccc}
    \toprule
    method     & backbone     & data &      \#epoch     &  mIoU  \\
    \midrule
    MAE  & ViT-S  & CC3M &  400        &    40.7      \\
    \OURS~(ours)  & ViT-S  & CC3M &  400        &    \textbf{43.3}      \\
    \midrule
    CLIP  & ViT-B  & CC3M &  400        &    39.4      \\
    SLIP  & ViT-B  & CC3M &  400        &    38.1      \\
    \color{fontgray}MoCoV3$*$  & \color{fontgray}ViT-B  & \color{fontgray}IN-1K &  \color{fontgray}1600        &    \color{fontgray}47.3      \\
    \color{fontgray}MAE$*$  & \color{fontgray}ViT-B  & \color{fontgray}IN-1K &  \color{fontgray}1600        &    \color{fontgray}48.1      \\
    MoCoV3  & ViT-B  & CC3M &  400        &    46.5      \\
    MAE  & ViT-B  & CC3M &  400        &    45.5      \\
    \rowcolor{mygray} \OURS~(ours)  & ViT-B  & CC3M &  400        &    \textbf{47.5}      \\
    \midrule
    MAE  & ViT-L  & CC3M &  400        &    50.0      \\
    \rowcolor{mygray} \OURS~(ours)  & ViT-L  & CC3M &  400        &    \textbf{52.1}      \\
    \bottomrule
    \end{tabular}
\end{table}

\noindent \textbf{Semantic segmentation on ADE20K.}
We transfer \OURS to a semantic segmentation task using the ADE20K dataset~\citep{ade20k}. 
Experiments on ADE20K use the decode head from UperNet~\citep{upernet}, and training recipes follow the default setting provided in \texttt{mmsegmentation}~\citep{mmseg2020}. 
Here, the training resolution is set to $512\times 512$, and the number of iterations is \texttt{160k}. 
We evaluate models with different capacities in~\cref{tab:ade20k_ft_main_result}. 
We observe that our \OURS approach results in significantly improved transfer performance across the ViT-S, ViT-B, and ViT-L backbones. 
We can see that the ViT-S MAE model only achieves 40.7\% mIoU. 
Consistent, with the previous IN-1K linear probing results, \OURS significantly alleviates the performance drop on the small backbone by improving mIoU by 2.6\%.
Notably when using the larger ViT-L backbone, \OURS  still improves over MAE by 2.1\% mIoU.
Compared to CLIP and MoCoV3, the superiority of \OURS  supports our information-theoretical motivation, \ie, generative pre-training results in a better transferability upper bound compared to discriminative pre-training. Consequently, we observe that generative pre-training greatly benefits downstream fine-tuned performance.

\subsection{Ablation Experiments}

\noindent \textbf{Ablation: Impact of image reconstruction and text generation losses. }
The most important ablation is the trade-off between the image reconstruction loss weight $\lambda_{V}$ and the caption generation loss $\lambda_{L}$. 
We fix $\lambda_{V}$ and ablate the value of $\lambda_{L}$. 
IN-1K fine-tuning experimental results are reported in~\cref{tab:ab_loss_weights}.  
The best transfer performance is achieved when $\lambda_{L}=0.1$, and both larger and smaller values perform worse.
This illustrates the trade-off between image reconstruction and text generation.  

Setting $\lambda_{L}=0.0$ makes \OURS equivalent to MAE. %
Increasing $\lambda_{L}$ in the range of (0.05, 0.3] increases the transfer performance, illustrating the effectiveness of our introduced text generation task. 
The reasons for the performance drop are two-fold:~(i)~As the CC3M pre-training data is web sourced, increasing $\lambda_{L}$ emphasizes the noisy text generation task. 
This can be problematic as captions coupled with web images can be often irrelevant, \eg, an image contains a river, but its caption may be about hiking. 
(ii)~As discussed in~\cref{sec:discussion}, the information contained in the text modality is greatly limited. Increasing $\lambda_{L}$ unavoidably leads to  over-fitting, which is unfavorable in representation learning. 
Therefore, a trade-off between image reconstruction and text generation exists.

We further study the effects of the image reconstruction target. 
We simply set $\lambda_{V}$ to $0.0$ and set $\lambda_{L}=1.0$, which only permits the text generation task. Interestingly, setting $\lambda_{V}$ to $0.0$ degrades \OURS to our MAC baseline.
Results are reported in~\cref{tab:ab_loss_weights} (right), where a notable performance drop is observed. 
This validates that~(1)~the information in text modality alone is insufficient, and~(2)~motivates the need for image generation for effective visual representation learning with image-text pairs.

\begin{figure*}[t]
    \centering
    \includegraphics[width=1.\textwidth]{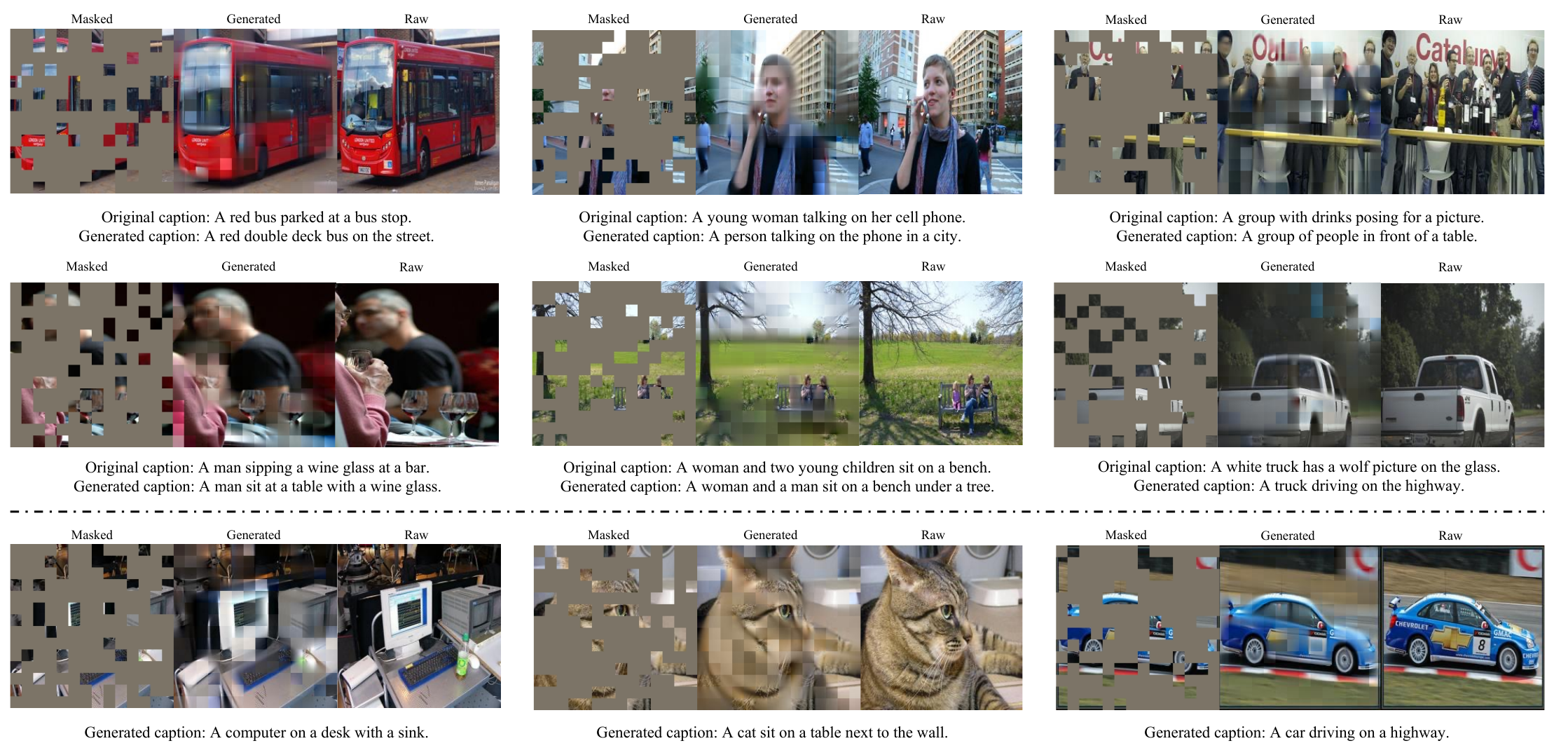}
    \caption{Reconstructions of masked images and captions from our \OURS approach from the MS-COCO~(top) and PASCAL-VOC~(bottom) datasets.}
    \label{fig:reconstruction}
\end{figure*}

\begin{table}[t]
    \centering
    \small
    \caption{\textbf{\textit{Left:}} Ablation of the language generation loss weight for \OURS pre-trained on CC3M and fine-tuned on ImageNet-1k. \textbf{\textit{Right:}} If we disable the visual generation task, performance suffers. }
    \label{tab:ab_loss_weights}
    \begin{minipage}[c]{0.7\linewidth}
    \setlength{\tabcolsep}{1mm}{
    \begin{tabular}{l|ccccc}
    \hline
    $\lambda_{L}$    &  0.0  &  0.05  & 0.1 & 0.2 & 0.3 \\
    \hline
    ft. top-1 acc.                             &  83.0   &  83.3   &  \textbf{83.5}   &  83.2  &  83.0 \\
    \hline
    \end{tabular}}
    \end{minipage}
    \begin{minipage}[c]{0.25\linewidth}
    \setlength{\tabcolsep}{1mm}{
    \begin{tabular}{cc}
    \hline
    w/o~$\mathcal{L}_{V}$ &  w/~$\mathcal{L}_{V}$ \\
    \hline
    81.7   &  \textbf{83.5}    \\
    \hline
    \end{tabular}}
    \end{minipage}
\end{table}

\noindent \textbf{Ablation: Impact of the number of layers in text encoder. }
For the text generation task, both single-modal and multi-modal layers are involved. 
We set the total number of layers to ten, and perform a grid search for the best result, see~\cref{tab:ab_textual_layer}. 
We observe that setting the number of single-layers to one or two achieves comparable performance. 
However, when the number of multi-layers decreases to five and six, the performance drop is significant. 
The reasons are likely two-fold: (i)~A few single-modal layers can support the text generation task since the single-modal layers can serve as a key entity word extractor, which only has a small influence on visual representation learning. 
(ii)~A certain number of multi-modal layers could be required to support the text generation task. An inadequate number of multi-modal layers further hurts the learning of image reconstruction because the performances becomes worse than the MAE baseline~(83.0\%).

\begin{table}[h]
    \centering
    \small
    \caption{Ablation of the number of textual layers for \OURS when pre-trained on CC3M and fine-tuned on ImageNet-1k.}
    \label{tab:ab_textual_layer}
    \begin{tabular}{l|ccccc}
    \hline
    \#single-/\#multi-layers    &  1/9  &  2/8  & 3/7 & 4/6 & 5/5 \\
    \hline
    F.T. top-1 acc.                             &  83.3   &  \textbf{83.5}   &  83.0   &  82.5  &  82.1 \\
    \hline
    \end{tabular}
\end{table}

\noindent \textbf{Ablation: Impact of form of text generation. }
Here we compare different text generation methods. 
There are two methods for text generation, \ie, masked language modeling~(MLM) and auto-regressive modeling~(captioning). 
We set the optimal $\lambda_{L}$ values for both methods, and observe that captioning outperforms MLM by 0.5\% top-1 accuracy~(83.5\% v.s. 83.0\%) on IN-1k. 
A harder task could help to approach a higher $I(X^V,X^L;T)$. 
Captioning is regularized to sequentially predict words, while MLM is not regularized. 
The text generator is composed of single-modal and cross-modal parts, and the text generation methods actually affects the learning mechanism of the single-modal part. 
Concretely, the auto-regressive learning target  forces the single-modal encoder to extract key words, \eg, the word after an article could be a key entity noun. 
For masked language modeling, contextual words are possibly masked and predicted. 
However, such noisy and potentially meaningless reconstruction tasks likely do not benefit the vision encoder as much.

\noindent \textbf{Visualization: Reconstructed images with coupled text. }
In~\cref{fig:reconstruction}, we provide generated images and captions produced by \OURS on the MS-COCO~\citep{coco} and PASCAL-VOC datasets~\citep{pascal}. %
The input to \OURS only consists of a masked image and a start word~(\eg, ``a''). As observed, \OURS can recover the main objects and entities in the raw images and annotated captions, even if most of the image pixels are masked. 
It is reasonable that \OURS only describes visible pixels, \eg, in the middle of the second row, the left child is masked, and thus \OURS only describes ``a woman and a man''. 
These visualizations show that the representation learned by \OURS can generate the joint distribution $p(X^V, X^L)$.

\section{Conclusion}
In this work, we first benchmarked different self-supervised representation learning methods and outlined two main observations: (i)~generative pre-training achieves the best transfer performance on web sourced datasets, and (ii) current multi-modal methods do not outperform single-modal ones.  
An information-theoretical view was developed to understand these benchmarking results and our observations. 
This inspired us to propose a new vision learner, titled MUlti-modal Generator~(MUG), that makes use of web sourced image-text paired data.
\OURS learns visual latent representations using two generative objectives and demonstrates strong generalizability across multiple transfer learning tasks with satisfactory scaling performance, and validates our information-theoretical perspective.

{
\bibliographystyle{tmlr}
\bibliography{egbib}
}

\clearpage
\newpage

\appendix

\section{Appendix}

\subsection{Implementation Details}

\noindent{\bf Pre-training.}
The default setting is in~\cref{tab:impl_mae_pretrain}, and hyper-parameters mainly follow~\citet{MAE} for fair comparisons. For each caption, we set the max length to 70 and use a percentage of 20\% input words for processing. For each word, we mask it, replace it with a random word, or delete it with a probability of 50\%, 10\%, and 40\%, respectively. 

\begin{wraptable}[12]{l}{.45\linewidth}
\center
\scriptsize
\caption{Pre-training settings.}
\label{tab:impl_mae_pretrain} %
\begin{tabular}{l|l}
\toprule
config & value \\
\midrule
optimizer & AdamW \\
base learning rate & 1.5e-4 \\
weight decay & 0.05 \\
optimizer momentum & $\beta_1, \beta_2{=}0.9, 0.95$  \\
batch size & 4096 \\
learning rate schedule & cosine decay  \\
warmup epochs  & 40 \\
image augmentation & RandomResizedCrop \\
text augmentation & mask/replace/delete \\
$\lambda_{V}$ & 1.0 \\
$\lambda_{L}$ & 0.1 \\
\bottomrule
\end{tabular}
\end{wraptable}

\noindent{\bf End-to-end fine-tuning on IN-1K, iNat-17, iNat-18, and P365.}
We fine-tune models with the widely used recipe~\citet{MAE,beit} in~\cref{tab:impl_mae_finetune} except for the layer-wise decay rate is set to 0.7. On all evaluated datasets and pre-training models, we use the same setting for fair comparisons.

\noindent{\bf Linear probing.} Settings are provided in~\cref{tab:impl_mae_linear}. We follow~\citet{MAE} to introduce an extra BatchNorm layer without affine transformation for improving linear probing performances. For fair comparisons, all evaluated models are equipped with the extra layer.

\noindent{\bf Semantic segmentation on ADE20K.}
The ADE20K fine-tuning setting can be found in this \href{https://github.com/open-mmlab/mmsegmentation/blob/master/configs/mae/upernet_mae-base_fp16_8x2_512x512_160k_ade20k.py}{website}. We advise settings for ViT-S and ViT-L by following~\citet{deit} and~\citet{beit}, respectively. All evaluated pre-training models are fine-tuned with the same setting, and thus comparisons are fair.

\begin{table}[ht]
\parbox{0.45\linewidth}{
\center
\scriptsize
\caption{End-to-end fine-tuning settings.}
\label{tab:impl_mae_finetune} %
\begin{tabular}{l|l}
\toprule
config & value \\
\midrule
optimizer & AdamW \\
base learning rate & 1e-3 \\
weight decay & 0.05 \\
optimizer momentum & $\beta_1, \beta_2{=}0.9, 0.999$ \\
layer-wise lr decay  & 0.7 \\
batch size & 1024 \\
learning rate schedule & cosine decay \\
warmup epochs & 5 \\
training epochs & 100 (B), 50 (L) \\
augmentation & RandAug (9, 0.5)  \\
label smoothing  & 0.1 \\
mixup  & 0.8 \\
cutmix  & 1.0 \\
drop path  & 0.1  \\
\bottomrule
\end{tabular}
}
\hfill
\parbox{0.45\linewidth}{
\center
\scriptsize
\caption{Linear probing settings.} 
\label{tab:impl_mae_linear}
\begin{tabular}{l|l}
\toprule
config & value \\
\midrule
optimizer & LARS  \\
base learning rate & 0.1 \\
weight decay & 0 \\
optimizer momentum & 0.9 \\
batch size & 16384 \\
learning rate schedule & cosine decay \\
warmup epochs & 10 \\
training epochs & 90 \\
augmentation & RandomResizedCrop \\
\bottomrule
\end{tabular}
}
\end{table}

\subsection{Transfer Learning with Visual Prompts}
Recent progresses in transfer learning in the visual domain has demonstrated the effectiveness of using visual prompt tuning (VPT)~\citep{jia2022vpt}.
By adding a small set of learnable tokens to the visual token sequence, VPT~\citep{jia2022vpt} can match, or even outperform, the performance of fully end-to-end fine-tuning of the pre-trained model.
Thus VPT has been seen as an effective method for tuning a pre-trained model.
In this section, we demonstrate that \OURS learned representations are effective for use by VPT by evaluating on several fine-grained benchmarks from the original VPT paper~\citep{jia2022vpt}.
The results are presented in~\cref{tab:vpt_result}.
We can see that \OURS improves the performance on all the data that we tested, indicating that the features learned by \OURS are not only effective for full fine-tuning, but also effective for parameter efficient tuning, which is more practical in real applications.

\begin{table}[ht]
    \centering
    \small
    \setlength{\tabcolsep}{4pt}
    \caption{
    Comparison classification performance using visual prompt tuning~\citep{jia2022vpt} across different pre-training methods and datasets. 
    }
    \label{tab:vpt_result}
    \begin{tabular}{lccccc}
    \toprule
    method   & data  & \#epoch     & CUB & SCars  & NABirds \\
    \midrule
    \color{fontgray}MAE$*$     & \color{fontgray}IN-1K &  \color{fontgray}1600  & \color{fontgray}68.4 & \color{fontgray}67.7 & \color{fontgray}65.2      \\
    MAE     & CC3M &  400                & 69.7    & 68.8  & 67.1     \\
    CLIP    & CC3M &  400                & 67.8    & 68.4  & 66.3     \\
    \rowcolor{mygray} 
    MUG (ours)  & CC3M & 400              & \textbf{70.1} & 69.3  & 67.4        \\
    \midrule
    MAE     & W200M &  10                & 70.2    & 69.0  & 67.2     \\
    CLIP    & W200M &  10                & 69.0    & 68.7  & 66.1     \\
    \rowcolor{mygray} 
    MUG (ours)  & W200M & 10              & \textbf{70.5} & 69.5  & 68.0        \\
    \midrule
    MAE    &  LAION400M  & 5             & 70.4           & 69.6  & 68.1  \\
    CLIP   &  LAION400M  & 5             & 69.8           & 68.9  & 67.3  \\
    \rowcolor{mygray}
    MUG (ours) & LAION400M & 5            & \textbf{70.6}  & 69.7  & 68.8   \\
    \bottomrule
    \end{tabular}
\end{table}

\subsection{More Diverse Evaluation}
In this section, we provide linear probing results and visual prompt tuning results on additional datasets.
These results complement the probing results in~\cref{tab:in1k_lp_main_result} and prompt tuning results in~\cref{tab:vpt_result}. 
We can see that in terms of linear probing, both MAE and MUG do not achieve a good performance compared to other methods. 
However, in the visual prompt tuning experiments, MUG demonstrates strong performance. 
As demonstrated in the original MAE paper, linear probing results are not always correlated with the final transfer performance of the learned representations, as linear probing misses the opportunity to make use of strong, but non-linear, features. 
Thus, we are more interested in the visual prompt tuning results as they better demonstrate the learned representation's quality. The effective rank experiments in~\cref{tab:rankme_in1k_result} also demonstrate that MUG can learn high quality representations.

\begin{table}[ht]
\centering
\label{tab:linear_add}
\caption{Linear probing results.}
\begin{tabular}{cccccccc}
\toprule
Linear Probing & Data & Food101 & CIFAR100 & Aircraft & DTD & Pets & Caltech 101 \\ 
\midrule
MoCoV3 & CC3M & 75.6 & 68.5 & 30.5 & 64.5 & 68.9 & 82.3 \\ 
CLIP & CC3M & 77.6 & 70.7 & 32.0 & 66.0 & 70.5 & 85.3 \\ 
SLIP & CC3M & 83.1 & 71.5 & 37.0 & 75.9 & 75.8 & 90.7 \\ 
\midrule
MAE & CC3M & 70.4 & 63.4 & 25.6 & 61.3 & 63.1 & 76.9 \\ 
\rowcolor{mygray}MUG & CC3M & 72.3 & 64.5 & 28.7 & 63.4 & 67.0 & 79.8 \\ 
\bottomrule
\end{tabular}
\end{table}

\begin{table}[ht]
\centering
\label{tab:prompt_add}
\caption{Visual prompt tuning results.}
\begin{tabular}{cccccccc}
\toprule
Visual Prompt Tuning & Data & Food101 & CIFAR100 & Aircraft & DTD & Pets & Caltech 101 \\ 
\midrule
MoCoV3 & CC3M & 81.4 & 76.7 & 37.6 & 71.2 & 84.2 & 86.1 \\ 
CLIP & CC3M & 83.5 & 77.5 & 38.9 & 74.4 & 85.6 & 88.3 \\ 
SLIP & CC3M & 85.6 & 77.9 & 39.4 & 76.1 & 86.3 & 90.4 \\ 
\midrule
MAE & CC3M & 85.1 & 76.8 & 39.0 & 75.3 & 84.7 & 88.7 \\ 
\rowcolor{mygray}MUG & CC3M & 86.7 & 78.4 & 41.3 & 77.4 & 85.3 & 91.9 \\ 
\bottomrule
\end{tabular}
\end{table}

\subsection{Assessing Transferability using RankMe}
\label{sec:rankme}
There have been a number of works that assess the performance of a pre-trained model on a downstream task prior to fine-tuning the model on the actual downstream dataset, \eg,~\citet{garrido2022rankme,agrawal2022alpha}. RankMe~\citep{garrido2022rankme} is one such method that assesses the performance of a model using the effective rank of the features extracted by the model on a downstream task. 
This can be viewed as measuring part of the mutual information $I(X;Z)$, as the rank of the features describe the variation of the features across dimensions. 
$I(X;Z)$ describes not only the dimension-wise variation of the feature, but also the magnitude of variation within each of the dimensions.

As the measure of mutual information in high dimensional space is difficult~\citep{belghazi2018mutual}, we leverage RankMe to show that \OURS can indeed learn a superior visual representation. 
Here, this is measured in terms of having higher effective rank compared to other methods on features extracted from the ImageNet-1k dataset.
The results are shown in~\cref{tab:rankme_in1k_result}.
Overall, we  observe that  discriminative methods like MoCoV3 and CLIP have a lower effective rank than generative methods like MAE and \OURS. 
Furthermore, \OURS has the highest effective rank,  indicating its ability to enlarge the information bottleneck of the model.

\begin{table}[ht]
    \centering
    \small
    \setlength{\tabcolsep}{4pt}
    \caption{
    Comparison of the effective rank of different pre-trained models on ImageNet-1k. Here, higher rank is better. 
    }
    \label{tab:rankme_in1k_result}
    \begin{tabular}{lccc}
    \toprule
    method   & data  & \#epoch     & Rank \\
    \midrule
    MoCoV3  & W200M &  10                & 413       \\
    MAE     & W200M &  10                & 674       \\
    CLIP  & W200M & 10                   & 312      \\
    \rowcolor{mygray} 
    MUG (ours)  & W200M & 10              & \textbf{704}       \\
    \midrule
    MAE    &  LAION400M  & 5             & 689  \\
    CLIP   &  LAION400M  & 5             & 425  \\
    \rowcolor{mygray}
    MUG (ours) & LAION400M & 5            & \textbf{792}   \\
    \bottomrule
    \end{tabular}

\end{table}

\subsection{Pre-training on Larger Datasets}

In this section we demonstrate the scability of \OURS by pre-training models on larger datasets such as LAION400M~\citep{laion400m} and W200M.
We mainly evaluate the performance of models using a ViT-B backbone.
For pre-training models on different sized datasets, we keep the total number of iterations the same across different datasets for a fair comparison.

\noindent \textbf{Fine-tuning on ImageNet-1k.}
We first compare the results of larger pre-training datasets for different pre-training methods on ImageNet-1k using end-to-end fine-tuning. 
Top-1 accuracy performance is reported in~\cref{tab:in1k_ft_data_result}.
We observe that similar trends from before still hold. Notably, \OURS achieves the best results across IN-1k, IN-A, and IN-R.
This demonstrate the scalability of \OURS, as the performance of \OURS still grows in line with the size of the pre-training dataset. 
In addition, \OURS is significantly better than other multi-modal baselines.

\begin{minipage}{1.0\linewidth}
\centering
\begin{minipage}{0.40\linewidth}
\centering
\begin{table}[H]
    \centering
    \small
    \setlength{\tabcolsep}{2pt}
    \caption{
    Comparison of methods on larger pre-training datasets on ImageNet-1k~(IN-1K) using end-to-end fine-tuning.
    }
    \label{tab:in1k_ft_data_result}
    \begin{tabular}{lccccc}
    \toprule
    method   & data  & \#epoch     & IN-1K & IN-A         &  IN-R  \\
    \midrule
    \color{fontgray}MAE$*$     & \color{fontgray}IN-1K &  \color{fontgray}1600  & \color{fontgray}83.6 &  \color{fontgray}35.9        &    \color{fontgray}48.3      \\
    MoCoV3  & W200M &  10                & 82.1 &  27.9 & 41.7      \\
    MAE     & W200M &  10                & 83.3 &  35.3 & 50.4      \\
    \midrule
    CLIP  & W200M & 10                   & 79.8 &  25.6 & 42.7      \\
    CoCa  & W200M & 10                   & 78.2 &  23.1 & 43.2      \\
    \rowcolor{mygray} 
    MUG (ours)  & W200M & 10              & \textbf{83.7} &  \textbf{39.0} & \textbf{51.8}      \\
    \midrule
    MAE    &  LAION400M  & 5             & 83.4  & 35.6 & 51.0 \\
    CLIP   &  LAION400M  & 5             & 80.1  & 29.7 & 45.2 \\
    \rowcolor{mygray}
    MUG (ours) & LAION400M & 5            & \textbf{83.9}  & \textbf{42.1} & \textbf{54.7} \\
    \bottomrule
    
    \end{tabular}
\end{table}
\end{minipage}
\hspace{0.10\linewidth}
\begin{minipage}{0.44\linewidth}
\begin{table}[H]
    \centering
    \small
    \setlength{\tabcolsep}{4pt}
    \caption{
    Comparison of different pre-training methods using larger datasets on ADE-20k using end-to-end semantic segmentation fine-tuning.
    }
    \label{tab:ade20k_ft_data_result}
    \begin{tabular}{lccc}
    \toprule
    method   & data  & \#epoch     & mIoU   \\
    \midrule
    \color{fontgray}MAE$*$     & \color{fontgray}IN-1K &  \color{fontgray}1600  & \color{fontgray}48.1      \\
    MoCoV3  & W200M &  10                & 46.0       \\
    MAE     & W200M &  10                & 46.3       \\
    \midrule
    CLIP  & W200M & 10                   & 40.3      \\
    \rowcolor{mygray} 
    MUG (ours)  & W200M & 10              & \textbf{48.5}       \\
    \midrule
    MAE    &  LAION400M  & 5             & 48.9  \\
    CLIP   &  LAION400M  & 5             & 43.2  \\
    \rowcolor{mygray}
    MUG (ours) & LAION400M & 5            & \textbf{50.0}   \\
    \bottomrule
    
    \end{tabular}
\end{table}
\end{minipage}
\end{minipage}

\noindent \textbf{Fine-tuning on ADE-20k.}
We also provide a comparison using larger pre-training datasets when fine-tuning for semantic segmentation downstream task on ADE-20k.
The results are presented in~\cref{tab:ade20k_ft_data_result}. 
It can be observed that \OURS can indeed improve the transfer learning performance as the pre-training dataset grows.
We note that \OURS achieves the best transfer learning results, again indicating the effectiveness of \OURS.

\subsection{Further Discussions on the Information Bottleneck View}
\label{app:sec:discussion}

In~\cref{sec:discussion}, we discussed about the inspirations of preliminary benchmark experiments. In this part, we extend the discussion from the following perspectives:

\noindent{\bf Why does the single/multi-modal discriminative method have a narrow bottleneck?}
Since popular discriminative methods are based on the contrastive learning, the nature of such methods could be regarded as the few-shot classification task, and the reason is as follows.
In each batch, we can regard each sample as a category, and the few-shot learning setting is B-way 1-shot, where B represents the batch size. For instance, in SimCLR~\citep{simclr}, an image is processed by two data augmentations as a positive pair, and ``pulling close'' the positive pair is actually minimizing a classification loss, \ie, the cross entropy loss. We believe the information compression rate of a classification task is relatively large, since the model only requires the most representative patterns to distinguish defined categories. Information irrelevant to the classification task will be compressed in the training process. Therefore, regarding discriminative methods as classification tasks would explain the narrow information bottleneck, and thus the $I(X;Z_d)$ would be relatively small because of significantly compressed input information. 
In addition, it also explains that contrastive learning pre-training methods require a large batch size~\citep{zerovl}, since increasing batch size will increase the difficulty of the few-shot classification task, which further alleviates the information compression. The narrow bottleneck also suggests that such methods require a certain amount of pre-training data, and insufficient data could result in the over-fitting problem.

\noindent{\bf Why does the single-modal generative method have a wide bottleneck?}
Popular single-modal generative methods are based on masked image modeling in an encoder-decoder fashion, and the working mechanism of such methods could be regarded as the image reconstruction task.
The image reconstruction task is much more challenging than the contrastive learning~(or few-shot classification) task, because the model is driven to preserve as much information as possible for recovering details~(concrete pixel values) in the raw input image. It explains the wide bottleneck, since the decoder cannot finish the reconstruction task without sufficient information of the encoder's output. Thus, the $I(X;Z_g)$ would be larger than $I(X;Z_d)$, and the upper bound is $H(X)$. Therefore, better transferring results are achieved~\citep{MAE,beit,beitv2} by using single-modal generative methods than discriminative methods.
The wide bottleneck also explains that generative methods can still perform well with a relatively small amount of pre-training data, as proven in~\citet{MAE}.

\noindent{\bf Why does the multi-modal generative method have a wider bottleneck?}
This work requires the model to reconstruct the raw input image and the coupled caption. It drives the model to preserve more information than the single-modal generative methods, since two objectives are involved and the decoder generates a joint distribution $p(X^V, X^L)$. The information bottleneck of a multi-modal generative method can be wider than that of a single-modal generative one.
In this way, we drive the $I(X;Z)$ to approach $H(X^V, X^L)$, which has the potential to learn more transferable representations.

\noindent{\bf How about scaling pre-training data?}
Note that the above discussions are based on a hypothesis that the pre-training data is fixed, \eg, CC3M in this work. Moreover, results in~\cref{sec:experiment} prove that the transferring results of our proposed multi-modal generative method are the best. We further discuss the choice of pre-training methods when the pre-training data is scalable. 

As aforementioned, single-modal discriminative methods could suffer from a large information compression rate; however, we can increase the amount of pre-training data as compensation. To prove this, we pre-train the SimCLR~\citep{simclr} with our collected W200M data, and then we transfer the model via ImageNet-1k end-to-end fine-tuning. MAE~(with W200M) and SimCLR~(with W200M) achieve 83.3\% and 83.4\% top-1 accuracies, respectively. It means that, with large-scale data, the drawback of single-modal discriminative methods could be overcome. Besides, it also suggests that the scalability of single-modal generative methods can be inferior. However, our method MUG has good scalability by achieving 83.7\%. 
Therefore, we suppose that our proposed method can perform well on different scales of pre-training data, which is an additional advantage.

\subsection{Additional Visualization Results}

\begin{figure*}
    \centering
    \includegraphics[width=\textwidth]{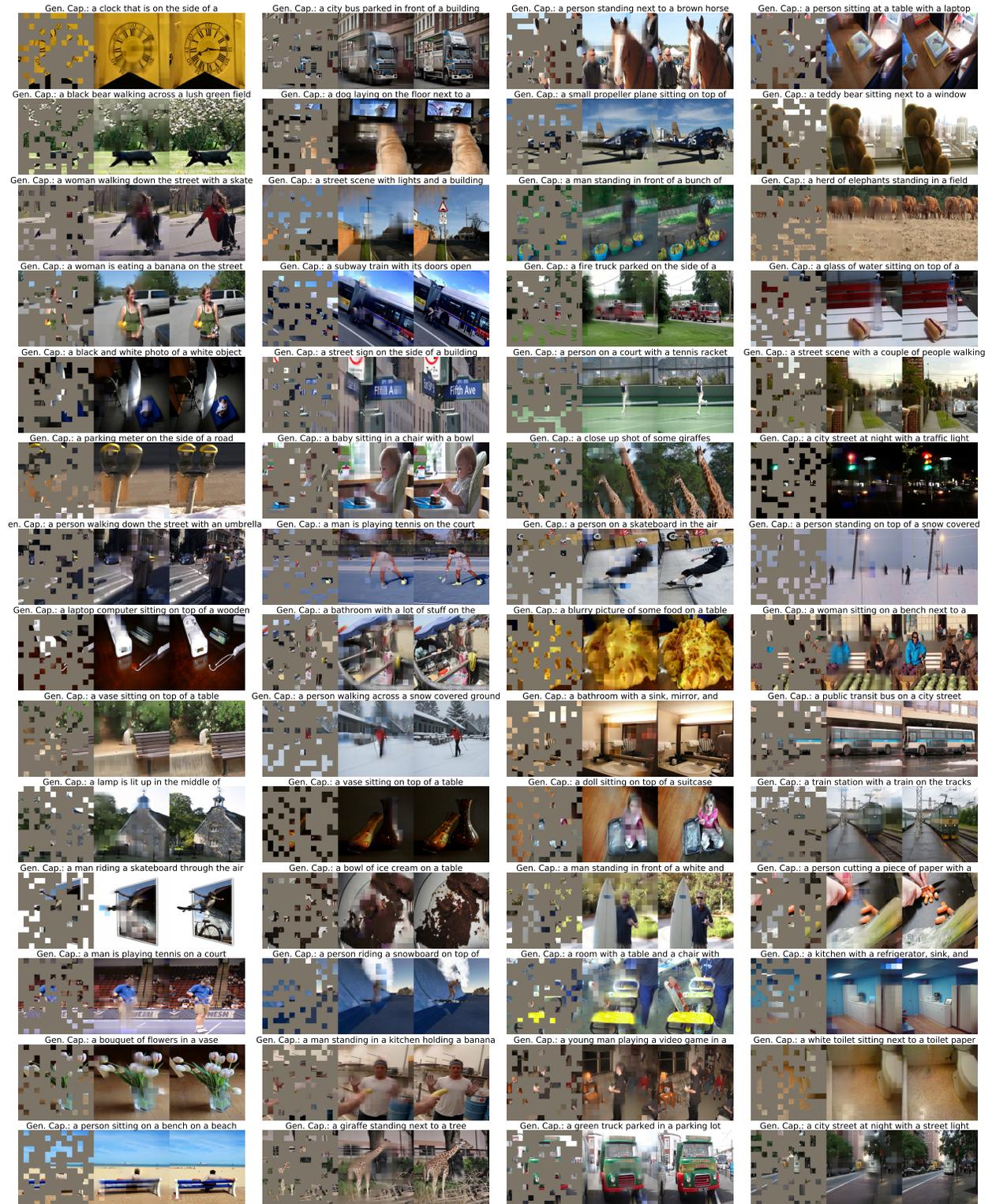}
    \caption{Uncurated random samples on COCO images. For each triplet, we show the masked image (left), \OURS reconstruction (middle), the ground-truth (right), and the generated captions by \OURS. The masking ratio is 75\%.}
    \label{fig:coco_random}
\end{figure*}

\begin{figure*}
    \centering
    \includegraphics[width=\textwidth]{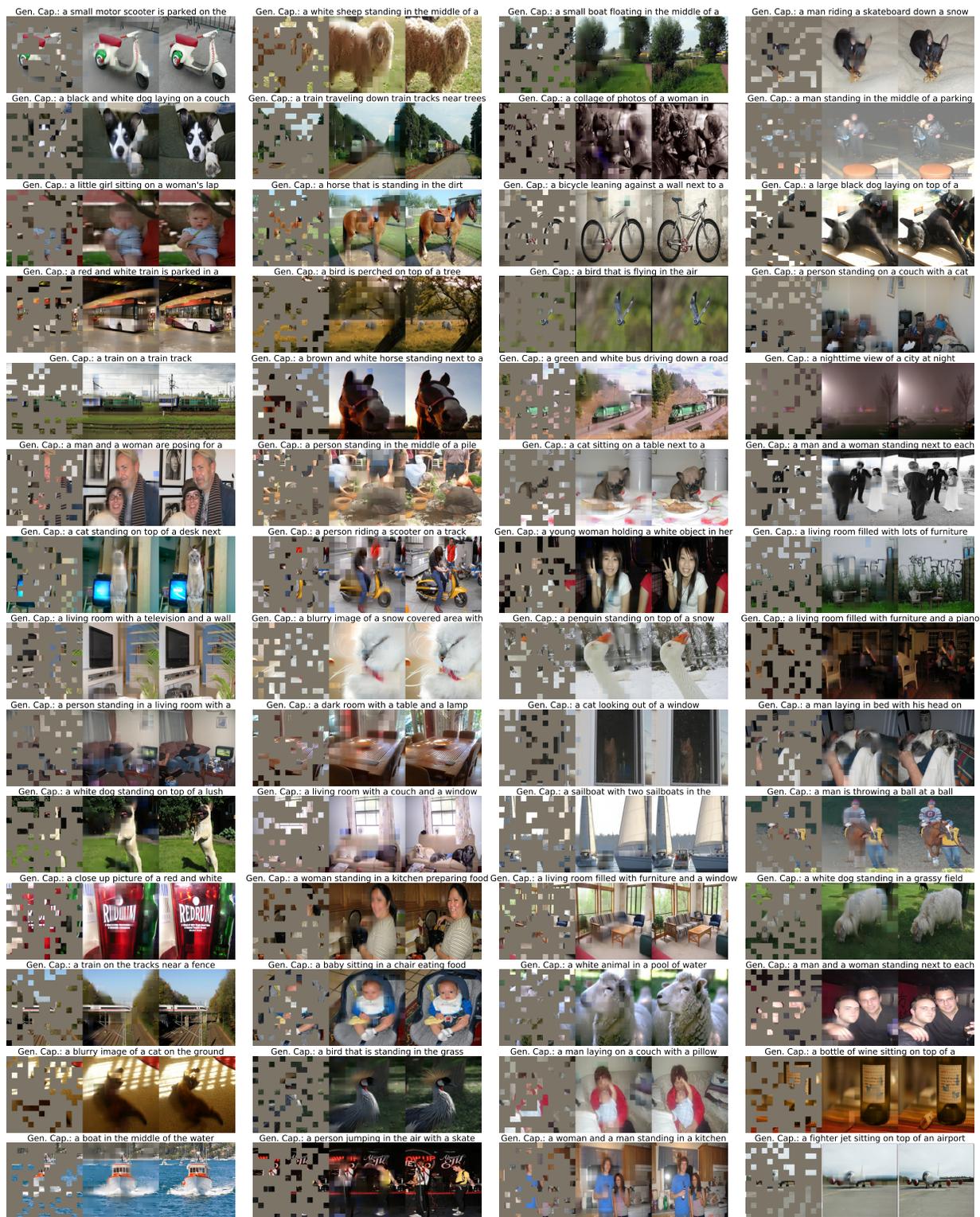}
    \caption{Uncurated random samples on PASCAL-VOC images. For each triplet, we show the masked image (left), \OURS reconstruction (middle), the ground-truth (right), and the generated captions by \OURS. The masking ratio is 75\%.}
    \label{fig:voc_random}
\end{figure*}

We further provide visualizations of the reconstructed images as well as the generated captions produced by our model on randomly selected images on the MS-COCO and PASCAL-VOC datasets in~\cref{fig:coco_random,fig:voc_random} for qualitatively understand the model's behaviour.

\subsection{Limitations and Border Impact}

One common limitation of models pre-trained on web sourced image-text pairs is that the models will inevitably be influenced by the open nature of internet data, such as biases or harmful speech. 
Thus the models may learn unwanted behaviors that may cause actual harm to people.
Given this limitation, resulting models should be tested thoroughly before they are deployed in real-world applications.

\end{document}